\documentclass[12pt,twoside]{article}   
\usepackage[super,sort,comma]{natbib}

\usepackage{comment}
\usepackage{graphicx}
\usepackage{amsmath}
\usepackage{amsfonts}
\usepackage[utf8]{inputenc}
\usepackage{ulem}

\makeatletter
\let\MYcaption\@makecaption
\makeatother

\usepackage{subcaption}
\makeatletter
\let\@makecaption\MYcaption
\makeatother

\newcommand{\beq}{\begin{eqnarray}}
\newcommand{\eeq}{\end{eqnarray}}

\usepackage{comment}
\usepackage{epstopdf}

\usepackage{fancyhdr}		

\newcommand{\captionv}[3]{\begin{center}\parbox{#1cm}{\caption[#2]{{\sf #3}}}
        \end{center}}

\usepackage[section]{placeins}   %

\usepackage{graphicx}

\makeatletter \renewcommand\@biblabel[1]{$^{#1}$} \makeatother
\newcommand{\cen}[1]{\begin{center} #1 \end{center}}

\usepackage{hyperref}
\hypersetup{ colorlinks,
    citecolor=blue,
    filecolor=blue,
    linkcolor=blue,
    urlcolor=blue
}

\usepackage{xcolor}

\definecolor{gray}{rgb}{0.6,0.6,0.6}
\definecolor{red}{rgb}{0.85,0,0}
\definecolor{green}{rgb}{0,0.85,0}
\definecolor{blue}{rgb}{0,0,0.85}
\definecolor{beige}{rgb}{0.92,0.87,0.78}
\usepackage[all]{hypcap}    

\usepackage[capitalize]{cleveref}
\crefname{section}{Section}{Secs.}
\Crefname{section}{Section}{Sections}
\Crefname{table}{Table}{Tables}
\crefname{table}{Tab.}{Tabs.}

\usepackage[symbol]{footmisc}

\begin{document}

\cen{\sf {\Large {\bfseries Training of deep cross-modality conversion models with a small dataset, and their application in megavoltage CT to kilovoltage CT conversion } \\  
\vspace*{10mm}
Sho Ozaki\footnote[2]{Ozaki and Kaji should be considered joint first authors.}$^{1}$, Shizuo Kaji$^{2}$, Kanabu Nawa$^{3}$, Toshikazu Imae$^{3}$, Atsushi Aoki$^{3}$, Takahiro Nakamoto$^{4}$, Takeshi Ohta$^{3}$, Yuki Nozawa$^{3}$, Hideomi Yamashita$^{3}$, Akihiro Haga$^{5}$, and Keiichi Nakagawa$^{1}$} \\
\vspace{5mm}
1. Graduate School of Medicine, University of Tokyo, Tokyo 113-8655, Japan \\
2. Institute of Mathematics for Industry, Kyushu University, 744 Motooka, Nishi-ku, Fukuoka 819-0395, Japan \\
3. Department of Radiology, University of Tokyo Hospital, Tokyo 113-8655, Japan \\
4. Department of Biological Science and Engineering, Faculty of Health Sciences, Hokkaido University, N12-W5, Kita-ku, Sapporo, Hokkaido, 060-0812, Japan \\
5. Graduate School of Biomedical Science, Tokushima University, Tokushima 770-8503, Japan
\vspace{5mm}\\
}

\pagenumbering{roman}
\setcounter{page}{1}
\pagestyle{plain}
Authors to whom correspondence should be addressed. \\
email:  shoozaki0117@gmail.com,
skaji@imi.kyushu-u.ac.jp\\

\begin{abstract}
\noindent {\bf Purpose:} In recent years, deep-learning-based image processing has emerged as a valuable tool for medical imaging owing to its high performance.
However, the quality of deep-learning-based methods heavily relies on the amount of training data; the high cost of acquiring a large dataset is a limitation to their utilization
in medical fields.
Herein, based on deep learning, we developed a computed tomography (CT) modality conversion method requiring only a few unsupervised images.\\
{\bf Methods:} The proposed method is based on CycleGAN with several extensions
tailored for CT images, which aims at
preserving the structure in the processed images and reducing the amount of training data.
This method was applied to realize the conversion of megavoltage computed tomography (MVCT) to kilovoltage computed tomography (kVCT) images.
Training was conducted using several datasets acquired from patients with head and neck cancer.
The size of the datasets ranged from 16 slices (two patients) to 2745 slices (137 patients) for MVCT and 2824 slices (98 patients) for kVCT.\\
{\bf Results:} 
The required size of the training data was found to be as small as a few hundred slices.
By statistical and visual evaluations,
the quality improvement and structure preservation
 of the MVCT images converted by the proposed model were investigated.
As a clinical benefit, it was observed by medical doctors that the converted images enhanced the precision of contouring. 
\\
{\bf Conclusions:} We developed an MVCT to kVCT conversion model based on deep learning, which can be trained using only a few hundred unpaired images.
The stability of the model against changes in data size was demonstrated.
This study promotes the reliable use of deep learning in clinical medicine by partially answering commonly asked questions, such as 
``Is our data sufficient?'' and ``How much data should we acquire?''\\
\\
\end{abstract}

{\it{keywords:}} Deep learning; Cross-modality conversion; Computed tomography; Training data reduction

\newpage     

\tableofcontents

\newpage

\setlength{\baselineskip}{0.7cm}      

\pagenumbering{arabic}
\setcounter{page}{1}
\pagestyle{fancy}
\section{Introduction}
{
Megavoltage computed tomography (MVCT) is used in helical tomotherapy, which is an innovative technique to administer intensity-modulated radiotherapy (IMRT) for image-guided radiotherapy (IGRT)~\cite{Mackie:1993, Mackie:2006, Meeks:2005, Ruchala:1999}. 
For precise registration based on image guidance, the quality of MVCT images must be ensured.
However, 
the image quality of MVCT is considerably lower than that of kilovoltage computed tomography (kVCT)
and thereby limits the accuracy of IGRT and adaptive radiotherapy (ART).
High-dose MVCT could enhance the image quality of MVCT~\cite{Westerly}.
However, increasing the dose inevitably leads to an increase in the patient's exposure to radiation. 
Therefore, it is desirable to develop a method that improves the image quality with the same scanning
data. This would also enable an improvement in the quality of images acquired in the past.
For this purpose, two main methods exist: reconstruction algorithms and image post-processing.
Iterative reconstruction with total variation regularization improves the image quality; however,
it is considerably slower than filtered back-projection
algorithms.
An accelerated, GPU-based iterative reconstruction scheme for MVCT has been developed previously ~\cite{ozaki:2020};
however, it requires approximately 10 min to reconstruct a single volume.

Image post-processing approaches include conventional denoising and deep-learning image processing techniques,
which generally offer good image quality within a short computational time.
In particular, deep learning has enabled image translation 
more than image quality enhancement.
In recent years, cross-modality conversion associated with different imaging modalities has gained considerable attention in the radiotherapy (RT) domain \cite{Spadea:2021, Wang:2021, kaji:2019a}.
In particular, the application of magnetic resonance imaging (MRI) to computed tomography (CT) conversion has been widely investigated in the context of MRI-based and MRI-guided RT~\cite{Wolterink:2017, Maspero:2018}.
Cone-beam computed tomography (CBCT) to CT conversion has been adopted to realize a higher accuracy of IGRT~\cite{kida:2019a} and facilitate its application to ART~\cite{Liang:2019, Taasti:2020}.
Recently, MVCT to kVCT conversion has also been investigated~\cite{Vinas:2021}.
In the conversion of CBCT (or MVCT) to kVCT,
the characteristics of the kVCT images are learned and transferred to CBCT (or MVCT) images to enhance the image quality.
These deep learning-based methods require considerable data, which comprise 
images in the corresponding modalities, for training. 
However, collecting a large dataset is inconvenient for patients and expensive for the medical staff.
In the aforementioned instances of modality conversions~\cite{Wolterink:2017, Maspero:2018, kida:2019a, Liang:2019, Taasti:2020, Vinas:2021}, the training data size ranged from 2795 to 6480 images.
Furthermore, a question often arises regarding the number of images
necessary to ensure successful training,
which is usually unknown until the model is trained.

To address this issue, in this study, we focused on reducing the number of training images required for cross-modality conversion to save the cost of data collection.
We developed a novel modality conversion model based on cycle-consistency generative adversarial networks (CycleGANs)~\cite{Zhu:2017} with several extensions, 
which performs well even when trained using a small dataset.
CycleGAN was originally proposed for natural image processing, 
where producing a diversified output was encouraged.
In other words, the structure in the image could be altered, 
which does not match the demand in the medical field.
Therefore, in this study, we introduced new loss functions designed specifically for the stability and preservation of these structures. 
In addition, certain data augmentation techniques, which we believed were suitable for our CT images, were applied.
The proposed model was applied in the MVCT to kVCT conversion of images of the mandible to neck regions,
where the original MVCT images were processed to produce kVCT-like images.
Its performance was evaluated
in terms of the histogram of Hounsfield Unit (HU) values, edge alignment
between the original MVCT and the processed images, and reduction of noise.
In particular, the dependence of these metrics on the size of the training dataset was closely investigated
to reveal that a limited number of 256 slices was sufficient
for successful training.
To highlight the clinical relevance, we conducted an evaluation in which medical doctors contoured a specific soft tissue in the processed and original MVCT images; based on the processed images, the
interpersonal variability in terms of the Dice coefficient was observed to be lower
than that with the original MVCT.
These findings can partially address the questions regarding the adequate number of training images for deep learning, which, in turn, can facilitate the reliable use of deep learning in clinical medicine.

The remainder of this paper is organized as follows.
\Cref{sec:method} describes the evaluation methods, data acquisition procedure, and
construction of the proposed model for modality conversion.
\Cref{sec:result} presents a comparison among MVCT, conventional denoising methods, reference kVCT, and processed MVCT images with visual and quantitative evaluations.
The dependence of the image quality and structure preservation
of the processed images on the size of the training data is emphasized.
To investigate the clinical benefit of the proposed 
method, the precision of soft tissue contouring 
by medical doctors was evaluated.
In \cref{sec:discussion}
a procedure to determine the necessary data size is discussed; in our specific case, the required number of images was found to be 256.
The generalizability of the model in converting images that are significantly different from the training images, including images with metal artifacts and other regions, was analyzed.
Lastly, \Cref{sec:conclusion} presents the concluding remarks.

\section{Materials and Methods}\label{sec:method}
\subsection{Data acquisition}\label{sec:data}
We used the MVCT and kVCT images of patients with head and neck cancer, who underwent IMRT via tomotherapy.
The MVCT images pertaining to tomotherapy (Accuray inc., US) scans were obtained using a helical fan-beam CT scanner with a tube voltage of $3.5$ MV, matrix size of $512 \times 512$ on the axial plane with a pixel size of $0.7462$ mm $\times$ $0.7462$ mm, and slice thickness of $3$ mm.
The kVCT images were obtained using a 16-row multidetector helical CT scanner of Aquilion LB (Canon Medical Systems, JP) with a tube voltage of $120$ kV, tube current of 350 mA, gantry rotation time of $0.5$ s, matrix size of $512 \times 512$ on the axial plane with a pixel size of $1.074$ mm $\times$ $1.074$ mm, and slice thickness of $2$ mm.
The MVCT and kVCT images were reconstructed through filtered back-projection (FBP).
The kVCT images, used as Planning CT (PlanCT), 
and the MVCT images for the image guidance
were acquired on different days
for each patient. 
Therefore, the kVCT images were not aligned with the MVCT images.
Because the pixel size of the kVCT images was different from that of the MVCT images, we upsampled the kVCT images to a pixel size of $0.7462$ mm $\times$ $0.7462$ mm, which was the same as that of the MVCT images.
To ensure training efficiency, the values of the CT images were clipped to [-600, 400] HU and scaled to [-1, 1].
In other words, pixels with an HU value below $-600$ and above $400$ were mapped to $-1$ and $1$, respectively. 

We used the slices from the mandible to the base of the neck,
which did not contain metal artifacts.
We acquired a total of 2745 slices from 137 patients for MVCT and 2824 slices from 98 patients for kVCT as the training data.
Moreover, we prepared an independent dataset comprising
319 MVCT and 488 kVCT images from a different group of 16 patients, referred to as \#1--\#16,
for validation.

\subsection{Modality conversion model trained with unpaired images}\label{sec:model}

\subsubsection*{Architecture of image translation model}
Our image conversion model comprises several neural networks whose 
architecture is based on popular unsupervised models such as CycleGAN \cite{Zhu:2017},
DiscoGAN \cite{discogan}, and DualGAN \cite{dualgan}. 
The general idea of generative adversarial network-based (GAN-based) image conversion can be found in
a previous survey~\cite{kaji:2019a}.
	
The proposed model uses an MVCT image slice, $x$, as the input and outputs a processed image slice, $y$, that resembles a kVCT image, and vice versa.
The model comprises three types of neural networks (\cref{Fig:networks_gen}). 
An \emph{encoder} is a network that, using an image as an input, outputs 
a latent vector, which is an abstract and high-level representation of the input image.
An encoder consists of down convolution layers and residual bottleneck layers.
A \emph{decoder} is a network that accepts a latent vector as the input and
produces an image;
this module comprises residual bottleneck layers and up-sampling layers.
The model includes two encoders, ${\rm{Enc}}_{\rm{MV \to latent}}$
and ${\rm{Enc}}_{\rm{kV \to latent}}$, and two decoders, 
${\rm{Dec}}_{\rm{latent \to kV}}$ and ${\rm{Dec}}_{\rm{latent \to MV}}$,
where ``latent'' indicates the latent vector space.
Certain combined networks (called generators) are formulated as follows:
\beq
G_{\rm{MV \to kV}}(x)
&=& {\rm{Dec}}_{\rm{latent \to kV}} ( {\rm{Enc}}_{\rm{MV \to latent}}(x) ), \\\
G_{\rm{kV \to MV}}(y)
&=& {\rm{Dec}}_{\rm{latent \to MV}} ( {\rm{Enc}}_{\rm{kV \to latent}}(y) ).
\eeq

The generator $G_{\rm{MV \to kV}}$ is trained to convert an MVCT image $x$
to a kVCT-like image $y=G_{\rm{MV \to kV}}(x)$;
$G_{\rm{{kV \to MV}}}$ is trained to convert a kVCT image $y$
to an MVCT-like image $x=G_{\rm{kV \to MV}}(y)$.
To ensure that the amount of information contained in the original and converted images
remains the same, the cycle consistency loss is defined as
\beq
\mathcal{L}_{\rm{cycle}}
&=& \sum_{x \in {\bf{MV}}} || x - G_{\rm{kV \to MV}} (G_{\rm{MV \to kV}}(x)) ||_{1}
+ \sum_{y \in {\bf{kV}}} || y - G_{\rm{MV \to kV}} ( G_{\rm{kV \to MV}}(y) ) ||_{1},
\eeq
which dictates that $G_{\rm{kV \to MV}}$ and $G_{\rm{MV \to kV}}$ are mutually inverse.
Here, $\bf{MV}$ and $\bf{kV}$ are the sets of training MVCT and kVCT images, respectively. 
With the cycle consistency loss alone, the generators would simply learn the 
identity map.
Thus, to learn the modality conversion, another element is necessary.

A \emph{discriminator} is a network that learns the distribution of 
images in a particular domain, in our case, MVCT or kVCT.
Its network structure is similar to that of the encoder, which
accepts an input image $x$ and outputs a feature vector $z$.
The only difference is that the discriminator is trained to 
output a vector $\mathbf{1}$ consisting of 1s if the input lies within the specified domain; otherwise, it outputs a vector $\mathbf{0}$ consisting of 0s.

The model includes two discriminators:
$D_{\rm{kV}}$, which is trained to learn the distribution of the real kVCT images to distinguish between real kVCT images and the processed MVCT (kVCT-like) images, and $D_{\rm{MV}}$, which is trained to learn the distribution of the real MVCT images to distinguish between the real MVCT images and the processed kVCT (MVCT-like) images.
These discriminators help the encoders and decoders to produce images 
according to the distribution of the real images.

To train $D_{\rm{kV}}$ and $D_{\rm{MV}}$, 
the loss function $\mathcal{L}_{D}$ for the discriminators is defined as follows:
\beq
\mathcal{L}_{D}
&=& \lambda_{D} \left\{ \sum_{x \in { {\bf{MV} } }} \left( || D_{\rm{kV}}( G_{\rm{MV \to kV}} (x) ) - {\bf{0}}  ||^{2}  
+ || D_{\rm{MV}}(x) - {\bf{1}} ||^{2} \right) \right. \nonumber \\
&& \left. + \sum_{y \in {\bf{kV}} } \left( || D_{\rm{MV}} (G_{\rm{kV \to MV}} (y) ) - {\bf{0}} ||^{2} + || D_{\rm{kV}} (y) - {\bf{1}} ||^{2} 
\right)  \right\}.
\label{Eq:Disc_loss}
\eeq

The generators and the discriminators are intertwined by
the adversarial loss, defined by
\beq
\mathcal{L}_{\rm{adv}}
&=& \sum_{x \in {\bf{MV}}} || D_{\rm{kV}} ( G_{ \rm{MV \to kV}} (x) ) - {\bf{1}} ||^{2} 
+ \sum_{y \in {\bf{kV}} } || D_{\rm{MV}} ( G_{\rm{kV \to MV}} (y) ) - {\bf{1}} ||^{2}. 
\eeq
This loss encourages the encoders and decoders to produce images according to the distribution
of the training data learned by the discriminators.

The elements described in this subsection are standard materials.

\subsubsection*{Techniques for improving stability and structure preservation}
\label{sec:technique}
We employed several extensions to the standard architecture described in the previous subsection.
Our extensions were tailored for the purpose of
achieving medical image conversion 
(a) that requires a reduced size of the training data
and (b) that preserves the anatomical structures.
Our extensions can be divided into the following categories: data augmentation and loss terms.

In addition to the random translation, we applied random rotation up to 20$^\circ$
and random scaling with a magnitude of 0.9--1.1, 
which were reasonable considering the variance in the patients' orientations on the couch
and body size.
Especially during the training, 
the images were randomly translated (up to 50 pixels), rotated, and scaled
before being cropped around the center to a size of $300 \times 360$ pixels.
This crop size was selected to ensure that the body always remained in the frame.

We added random Gaussian noise of $\sigma=0.01$ in the image domain and latent space.
This helped the generators to acquire a robust latent representation of the images.

\begin{figure*}[t]
\centering
\includegraphics[width=0.6\linewidth]{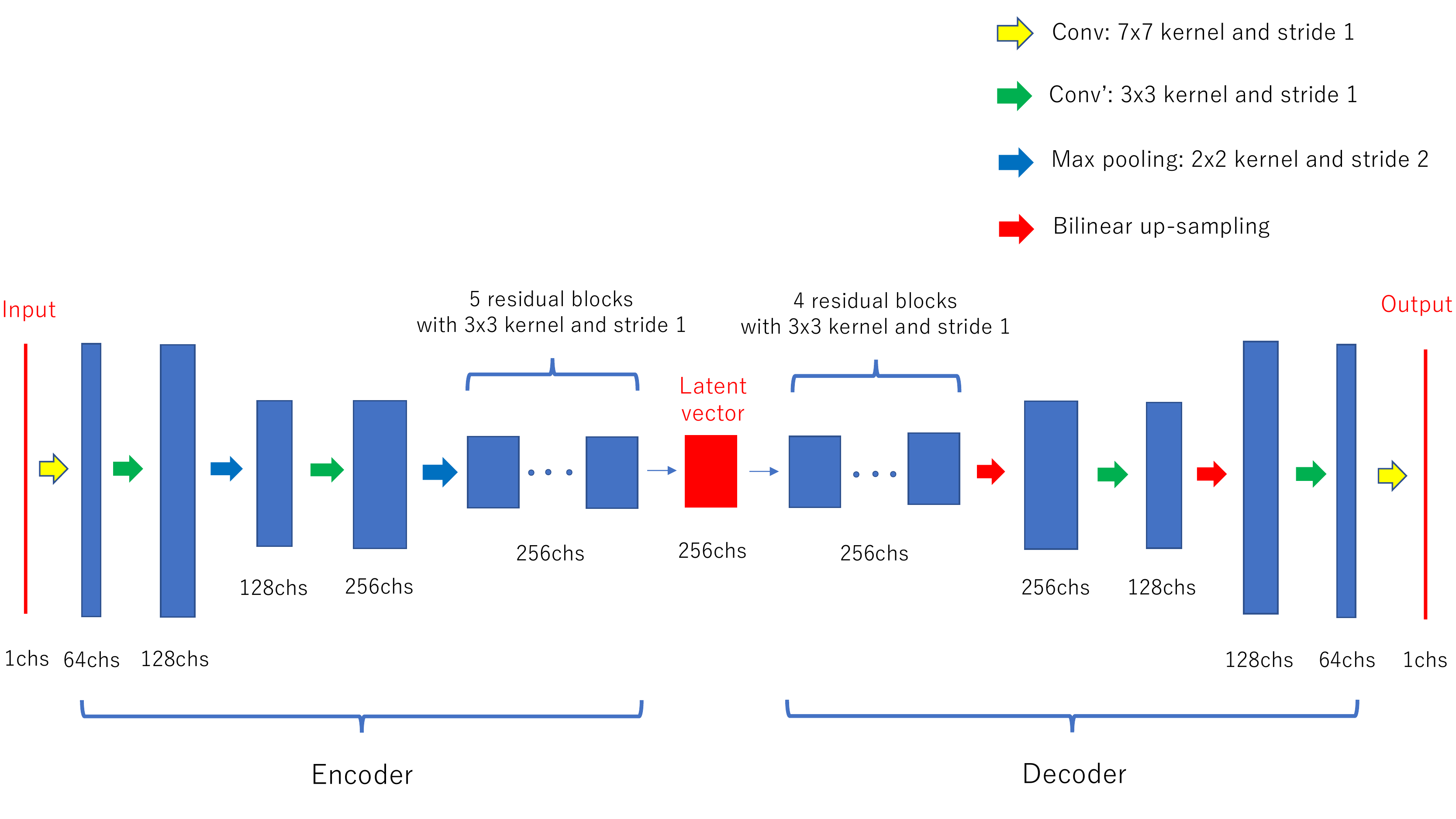}
\includegraphics[width=0.6\linewidth]{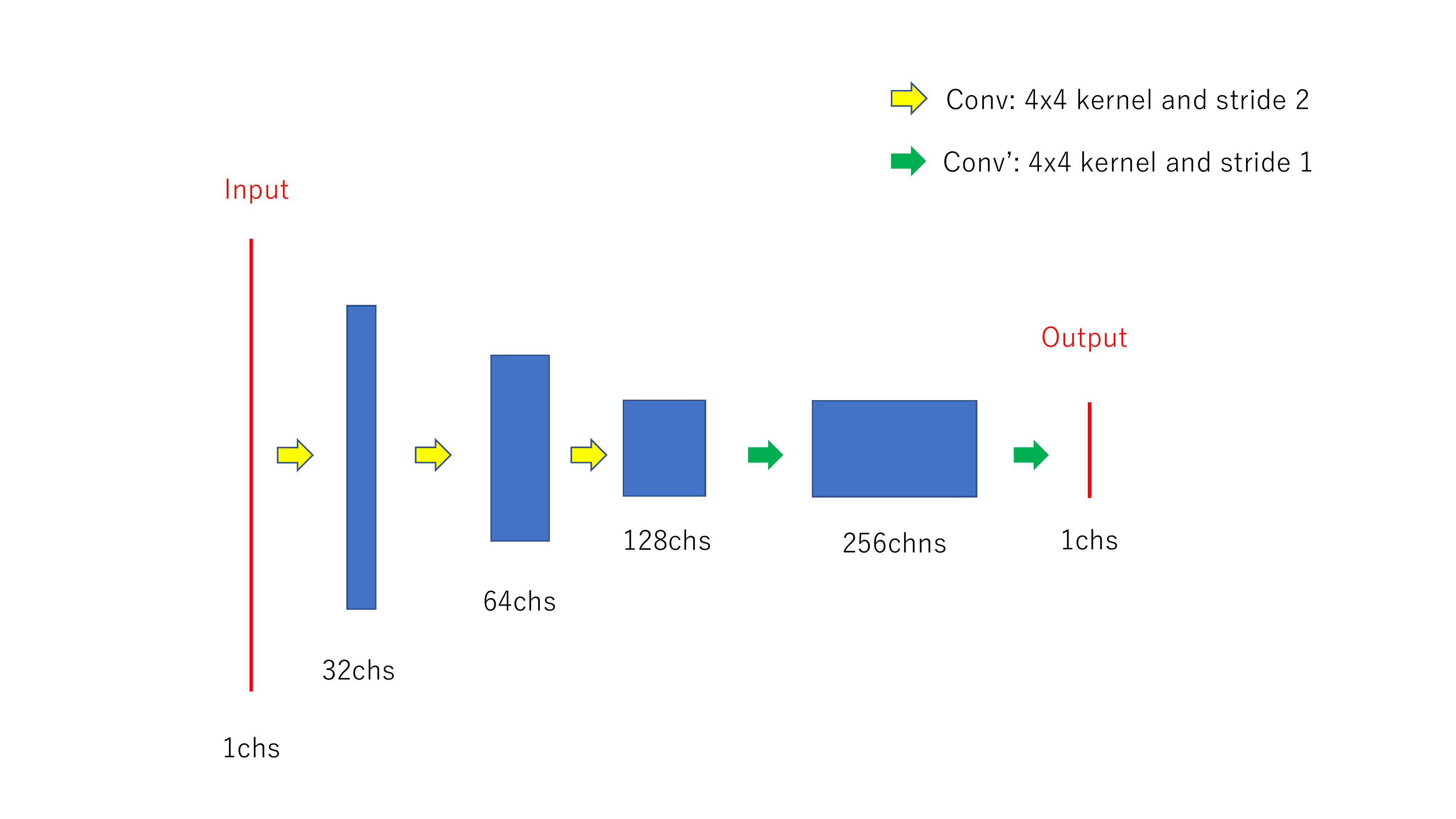}
\captionv{12}{}{Network architecture of generators
$G_{\rm{MV} \to \rm{kV}}$, $G_{\rm{kV} \to \rm{MV}}$, $G_{\rm{MV} \to \rm{MV}}$, $G_{\rm{kV} \to \rm{kV}}$
(top) and discriminators $D_{\rm{kV}}$, $D_{\rm{MV}}$
(bottom).
\label{Fig:networks_gen} }
\end{figure*}

The problem of image conversion is highly ill-posed; hence, adding loss terms generally contributes toward stability.
Moreover, to suppress the alteration of the anatomical structure
owing to the adversarial loss, we require an additional loss term.

The autoencoder loss is
\begin{align}
\mathcal{L}_{\rm{autoenc}}
=& \sum_{x \in {\bf{MV}}}  || x - {\rm{Dec}}_{\rm{latent \to MV}} ( {\rm{Enc}}_{\rm{MV \to latent}}(x) ) ||_{1} \nonumber \\
&+ \sum_{y \in {\bf{kV}}} || y - 
{\rm{Dec}}_{\rm{latent \to kV}} ( {\rm{Enc}}_{\rm{kV \to latent}}(y) )||_{1}.
\end{align}
The autoencoder loss strengthens 
the cycle consistency loss
to enforce the reversibility of the conversion, such 
that the information in the input image and its latent representation
 are equivalent.
The total variation loss is defined as
\beq
\mathcal{L}_{\rm{tv}}
&=& \sum_{x \in {\bf{MV}} } || {\rm{grad}} (G_{\rm{MV \to kV}} (x)) ||_{1},
\eeq
where grad is the image gradient.
Owing to this loss, the generator produces spatially uniform images.
The air-region loss is defined as
\beq
\mathcal{L}_{\rm{air}}
&=& \sum_{x \in {\bf{MV}}} || \psi(G_{\rm{MV \to kV}} (x)) - \psi(x) ||_{1} + \sum_{y \in {\bf{kV}}} || \psi(G_{\rm{kV \to MV}} (y)) - \psi(y) ||_{1},
\eeq
where
\beq
\psi(x)
&=& 
\begin{cases}
 x & ({\rm{if}} \ x < C ), \\
 0 & ({\rm{if}} \ x \ge C ),
 \end{cases}
\eeq
where $C$ is a constant equivalent to $-598$ HU.
This loss ensures that the generators do not alter 
the regions with values less than $-598$ HU to preserve the air--body boundaries.

The perceptual loss is defined as
\beq
\mathcal{L}_{\rm{percep}} 
&=& \frac{1}{whn_{\rm{cl}}} 
\left( \sum_{x \in {\bf{MV}} } || \phi( G_{\rm{MV \to kV}} (x) ) - \phi(x) ||^{2} + \sum_{y \in {\bf{kV}} } || \phi( G_{\rm{kV \to MV}} (y) ) - \phi(y) ||^{2} \right), \nonumber \\
\label{Eq:perceptual}
\eeq
where $\phi$ is the output of the second convolutional layer of the VGG16 networks pre-trained using the natural images derived from the ImageNet database, and
$w,h,n_{\rm{cl}}$ denote the width, height, and number of channels of this layer, respectively.
The second (and, in general, shallow)
layer of the VGG network is known to learn 
low-level image features such as edges.
\Cref{Fig:VGG} visualizes the channels of 
$\phi(x)$ for a slice $x$ from the MVCT of patient \#1.
They represent low-level image features, such as edges,
and the boundaries of different regions, and hence, the body contour is captured.
This suggests that maintaining a low perceptual loss facilitates 
the preservation of the structures between the input and output images.

\begin{figure*}[t]
\centering
\includegraphics[width=0.8 \linewidth]{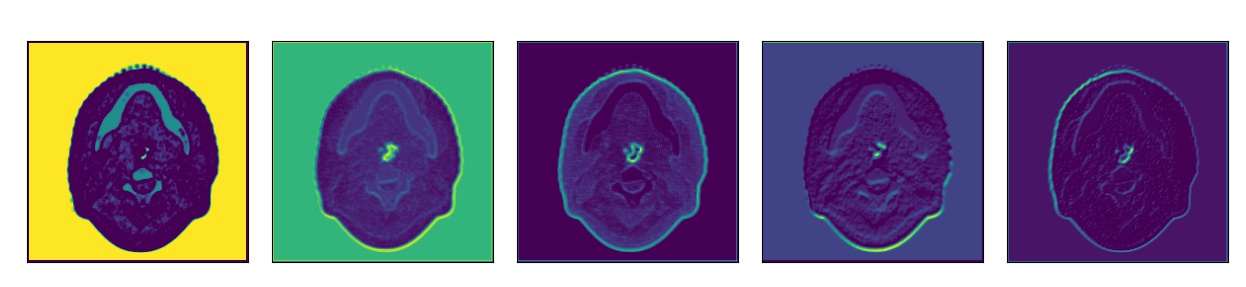}
\captionv{15}{}{
Example of the output of the second layer of VGG16
for an MVCT image of patient \# 1.
\label{Fig:VGG}
}
\end{figure*}

Consequently,
the loss function $\mathcal{L}_{G}$ for the encoders and decoders is defined by
\beq
\mathcal{L}_{G}
&=& \lambda_{\rm{cycle}} \mathcal{L}_{\rm{cycle}} + \lambda_{\rm{autoenc}} \mathcal{L}_{\rm{autoenc}} + \lambda_{\rm{adv}} \mathcal{L}_{\rm{adv}} + \lambda_{\rm{tv}} \mathcal{L}_{\rm{tv}}
+ \lambda_{\rm{air}} \mathcal{L}_{\rm{air}} + \lambda_{\rm{percep}} \mathcal{L}_{\rm{percep}},
\label{Eq:Loss_for_generators}
\eeq
where the hyperparameters were empirically selected as follows:
\beq
\lambda_{\rm{cycle}} = 10.0, \, \, \, \lambda_{\rm{autoenc}} = 1.0, \, \, \, \lambda_{\rm{adv}} = 0.1, \, \, \, \lambda_{\rm{tv}} = 0.001, \, \, \,  \lambda_{\rm{D}} = 1.0, \, \, \, \lambda_{\rm{air}} = 1.0, \, \, \, \lambda_{\rm{percep}} = 0.1. 
\label{Eq:Hyperparametes}
\eeq
A schema of the proposed networks and loss functions is presented in \cref{Fig:Schema_of_traning}.

\begin{figure*}[t]
\begin{tabular}{c}
\begin{minipage}{1.0\hsize}
\includegraphics[width=1.0 \textwidth, bb = 0 0 900 550]{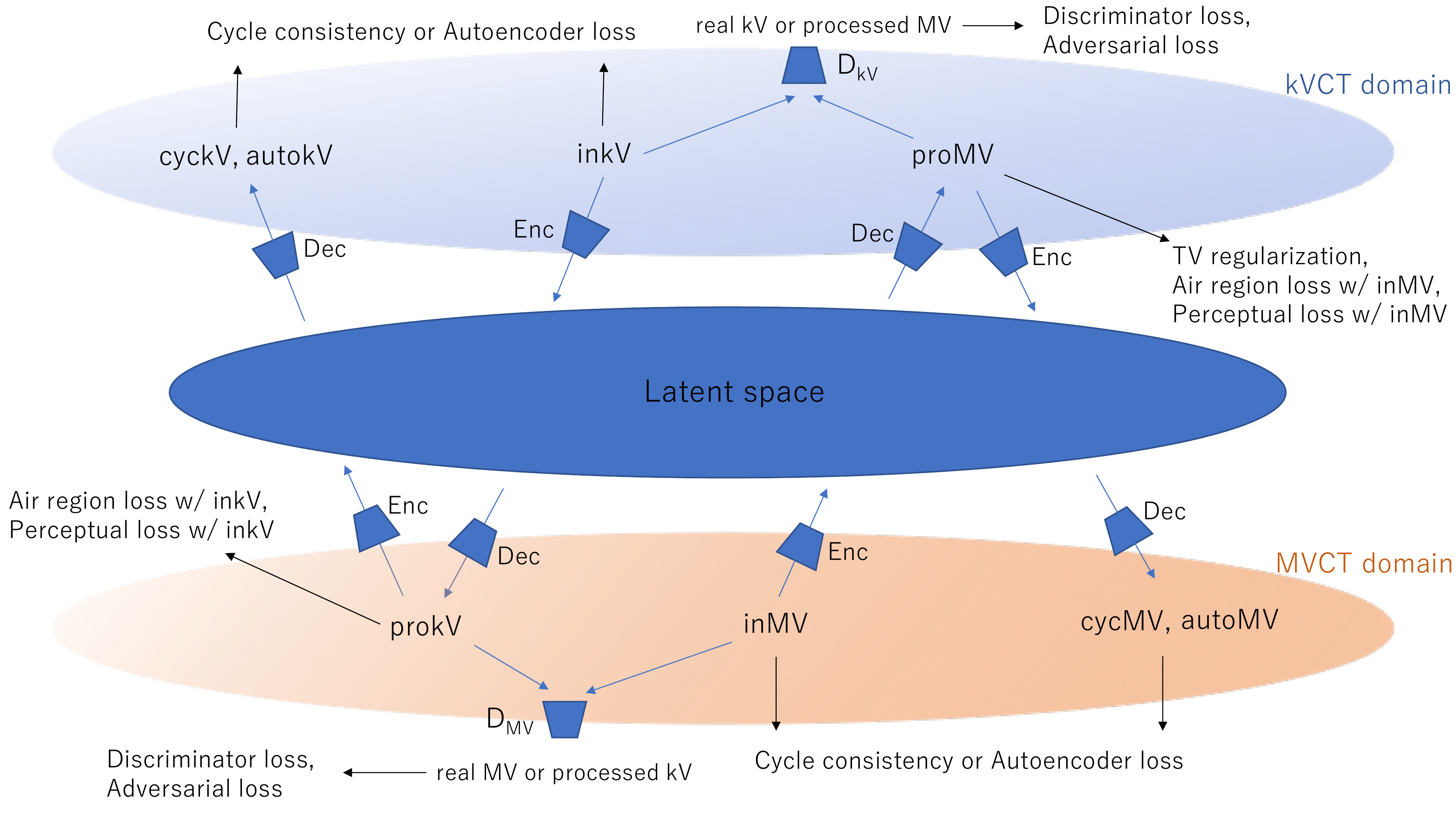}
\captionv{12}{}{Schema of our networks with loss functions.
\label{Fig:Schema_of_traning} }
\vspace{5mm}
\end{minipage}
\end{tabular}
\end{figure*}

In CBCT to kVCT conversion, a previous study~\cite{kida:2019a} used the following two loss
terms.
The idempotency loss function was defined by
\beq
\mathcal{L_{\rm{idem}}}
&=& \sum_{x \in {\bf{MV}}} || G_{\rm{MV \to kV}}(x) - G_{\rm{MV \to kV}} (G_{\rm{MV \to kV}}(x) ) ||_{1} \nonumber \\
&& + \sum_{y \in {\bf{kV}}} || G_{\rm{kV \to MV}}(y) - G_{\rm{kV \to MV}}(y)  ( G_{\rm{kV \to MV}}(y) ) ||_{1},
\eeq
and the gradient loss was defined as the $L^2$-norm
of the difference of the gradients concerning the input and output images.
The idempotency loss was derived from a formal requirement
for the mappings defined by generators and helped improve the stability. In the proposed model, 
the autoencoder loss plays a similar role and replaces the 
idempotency loss to reduce the computational cost.
The purpose of the gradient loss 
is to ensure the alignment of edges
between the input and output images by comparing them in the gradient domain.
This is superseded in the proposed model by the perceptual loss, which compares various low-level image features between the input and output images.

In the results section, we refer to the original CycleGAN for a comparison.
The original study~\cite{Zhu:2017}
adopted a loss function with
the cycle consistency loss, 
adversarial loss, and 
discriminator loss:
\begin{align}\label{eq:originalCycleGAN}
\mathcal{L}'_{G}
&= \lambda'_{\rm{cycle}} \mathcal{L}_{\rm{cycle}} + \lambda'_{\rm{adv}} \mathcal{L}_{\rm{adv}}, \ 
\mathcal{L}'_{D}=\mathcal{L}_{D},
\end{align}
where $\lambda'_{\rm{cycle}}=10.0$ and $\lambda'_{\rm{adv}} = 1.0$.
However, as we will see later in the Results section,
the resulting images are overly altered owing to the adversarial loss.
This can be mitigated by setting $\lambda'_{\rm{adv}}$ to a lower value.
For a fair comparison, we set $\lambda'_{\rm{adv}}= 0.1$ in our experiments when referring to
the \emph{original} CycleGAN for the comparison.
It was still observed that the proposed model performed better, which explains
the purpose of developing the extensions described in this section.

All the numerical experiments were conducted using a personal computer equipped with a single GPU (Nvidia 2080Ti) and CPU (Intel Core i9-9940X) with 64 GB memory and Ubuntu 16.04 LTS operating system.
We implemented our algorithm with Python 3.7.5 and Chainer 7.2.0. 
The codes used in this study are available on Github\footnote{https://github.com/shizuo-kaji/UnpairedImageTranslation}.
Training using all the training data (2745 slices of MVCT and 2824 slices of kVCT) for 100 epochs required approximately two days.
The batch size of 1 was employed, and the optimizer parameters were set to be the same as those for the original CycleGAN.
The conversion from MVCT images to processed MVCT images using the trained model
required approximately a few seconds for 20 slices of one patient.


\subsection{Quantitative evaluation by image metrics}\label{sec:metrics}
An independent dataset comprising 
 unaligned MVCT and kVCT images derived from sixteen patients with head and neck cancer, \#1--\#16, were used for the evaluation.
In addition to visual inspection 
and the statistical analysis on the histogram of HU values, we adopted the following evaluation metrics to quantify the validity of the proposed model.

To quantitatively evaluate the degree of noise reduction
in the processed MVCT (kVCT-like) images, we propose a new metric, noised SelfSSIM (NSelfSSIM), defined for an image $x$ as the mean
\beq
{\rm{NSelfSSIM}}(x) = E_{\epsilon \sim \mathcal{N}(0,\sigma^2)}[{\rm{SSIM}}(x, x + \epsilon)],
\label{Eq:NselfSSIM}
\eeq
where SSIM is the structural similarity index measure
and $\mathcal{N}(0,\sigma^2)$ is the zero-mean Gaussian distribution with $\sigma=20$ HU\footnote{
We choose $\sigma=20$ HU as the minimum strength of clinically relevant noise.
We also conducted tests with $\sigma=10$ HU and $\sigma=30$ HU.
The absolute values were different, but the relative behavior among different models
was observed to be similar, thereby indicating the robustness of the metric.}.
The rationale is that, 
if the image has low noise, 
it is significantly different in terms of the SSIM when Gaussian noise is added.
Accordingly, NSelfSSIM exhibits lower values
if the image has a lower noise level.
This metric can quantitatively indicate the noise reduction in the processed images.

To quantify structural preservation, 
we introduced the difference in gradient (DIG), which is defined as
\beq
\mathrm{DIG}(x) &=&  || {\rm{grad}}( x - G_{\rm{MV \to kV}} (x) ) ||^{2},
\label{Eq:DIG}
\eeq
where $x$ is the input MVCT image.
The DIG measures the change in
the pixel values of the original and 
processed images in the gradient domain,
thus facilitating an evaluation of the preservation of edges.

\subsection{Clinical evaluation}\label{sec:contour}
The clinical relevance of 
the quality improvement of the processed images was assessed through a contouring task.
Seven medical doctors (MDs) of radiotherapy from the University of Tokyo Hospital contoured the two parotid glands of the four patient images
in the validation dataset, including the processed MVCT by the proposed model, original MVCT, and PlanCT (kVCT) images.
In total, eight regions of interest (ROIs) were prepared under four different modalities and contoured by seven doctors.
To avoid the region of the metal artifact in the oral cavity, approximately 50 \% of the volume of each parotid was used.
We designed the experiments as follows.
The MDs were asked to contour the 1) MVCT of all four patients. 
After finishing the contouring, the images were taken away
and they were asked to contour 2) processed MVCT by the data-16 model 
(see \cref{sec:reduced_dataset}) of all four patients.
This procedure was repeated for 3) processed MVCT by the data-full model
and 4) PlanCT registered with MVCT.
Additionally, the same MDs were asked to contour 5) processed MVCT by the data-256 model
after an interval of more than one week,
which we believe is sufficiently long to nullify the effect of the doctors' memory.
For the evaluation,
we adopted the PlanCT (kVCT)
images acquired on the same day as the MVCT images for the same patient. 
3D rigid registration from the PlanCT images to the MVCT images was applied using a 3D Slicer~\cite{kikinsi}.
We considered the average of the seven contours 
on the registered PlanCT images
provided by the doctors and regarded the average contour as the ground truth.
The average of the contours refers to 
the binarization of the average of binary images at a threshold of 0.5.
We computed the Dice coefficients 
between the contours in the processed MVCT and the ground truth, and those between the
contours in the original MVCT and the ground truth.

\section{Results}\label{sec:result}

\subsection{Comparison with conventional denoising}

\Cref{Fig:VisualComp} shows
the MVCT, non-local means denoising (NLM), non-local means denoising with histogram matching (NLM+HM),
processed MVCT by the proposed model, and kVCT of the two selected slices of patients \#1 and \#2.
The patch size and the patch search area for the non-local means denoising were set to $5\times 5$ and $13\times 13$, respectively. The histogram of the NLM image was matched with that of the kVCT
for NLM+HM.
kVCT images of the same patients, acquired on different days, are shown for reference.
It should be noted that the processed MVCT image is not expected to accurately match the kVCT image because the MVCT and kVCT images were obtained on different days.
The figures highlight the enhancements in image quality afforded by the proposed model.
In particular, the tiny structure, which was visible in the kVCT but vague in the original and denoised MVCT (NLM and NLM+HM),
was clearly identified
in the processed MVCT by the proposed method 
owing to the improved soft tissue contrast.
A similar improvement was observed for all the validation datasets, i.e., \#1--\#16.

\begin{figure*}[t]
\centering
\includegraphics[width=0.88 \textwidth]{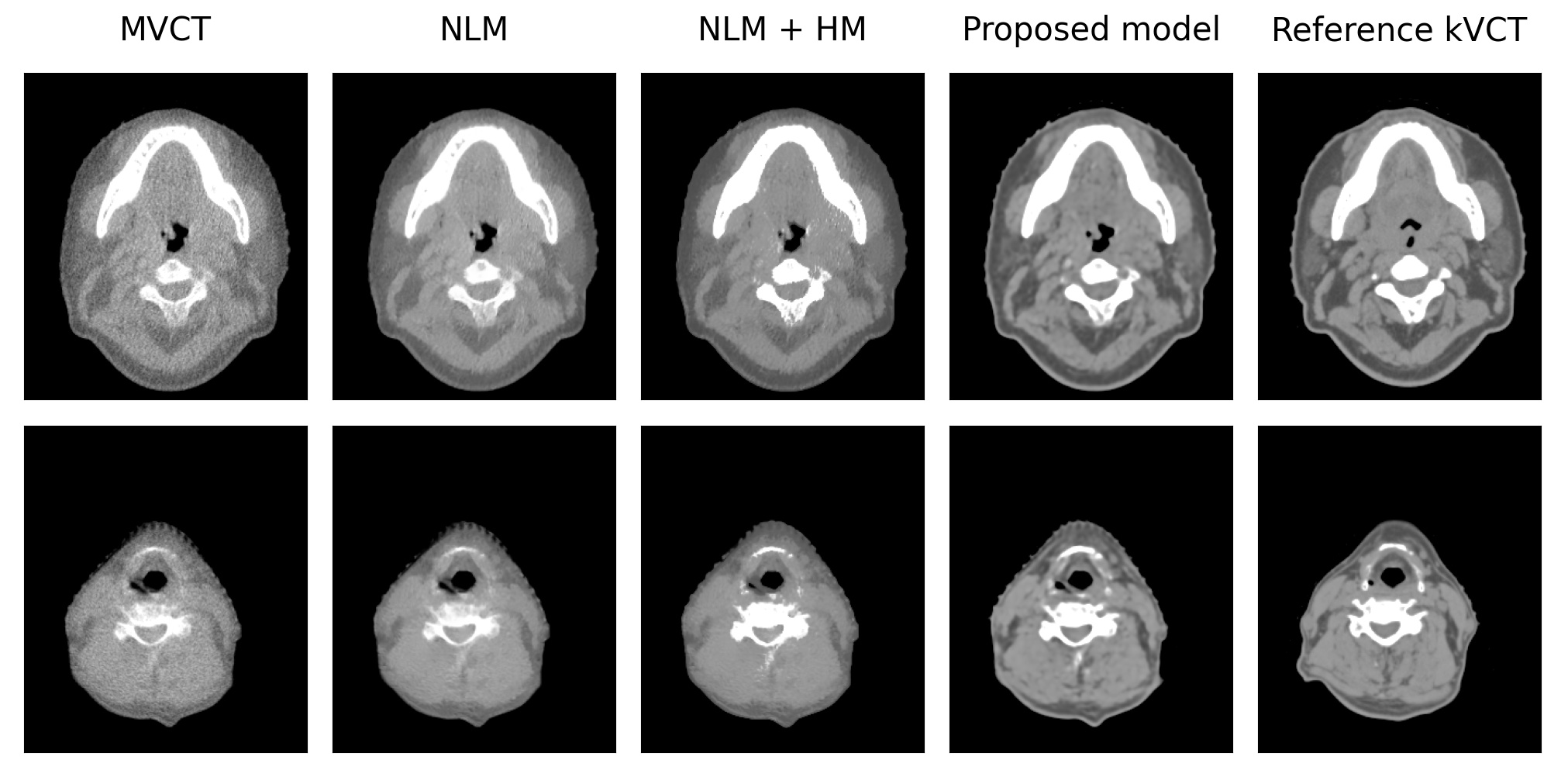}
\includegraphics[width=0.87 \textwidth]{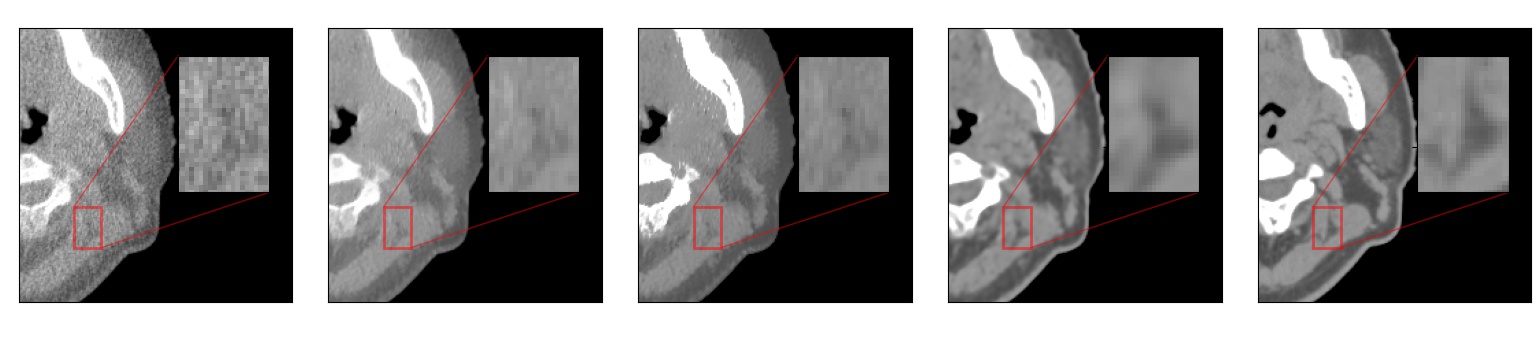}
\captionv{15}{}{
Visual comparison of image quality among MVCT, non-local means denoising (NLM), 
non-local means denoising with histogram matching (NLM+HM), processed MVCT by data-full, and reference kVCT images of patients \#1 (first row) and \#2 (second row).
A tiny structure in the muscle of patient \#1 is highlighted (third row).
The kVCT was acquired on a different day.
The display window range is set as (-300, 300) HU.
\label{Fig:VisualComp}
}
\end{figure*}

\subsection{Training the model with datasets of various sizes}\label{sec:reduced_dataset}

To estimate the required amount of data for 
MVCT to kVCT conversion with the proposed model,
we prepared small-sized datasets comprising 16, 32, 64, 128, 256, 512, 1024, and 2048 slices for each MVCT and kVCT case, labeled as data-16, data-32, data-64, data-128, data-256, data-512, data-1024, data-2048, and data-full, respectively,
as subsets of the training data described in \cref{sec:data}.
Data-full represents the dataset containing all the images (2745 slices of MVCT and 2824 slices of kVCT).
The number of patients and slices in each dataset are summarized in \cref{Table:Datastes}.
We refer to the proposed model trained with, for example, data-16, as the data-16 model.

\begin{table}[htbp]
\captionv{15}{}{
Number of patients and slices contained in the training datasets
and the number of epochs for the training.
\label{Table:Datastes}
}
 \centering
 \scalebox{0.68}{
  \begin{tabular}{lccccccccc}
  \hline \hline
   & data-16  & data-32 & data-64 & data-128 & data-256 & data-512 & data-1024 & data-2048 & data-full  \\
  \hline
  Patients & 2  & 4 & 8 & 16 & 32 & 64 & 80 & 120 (MVCT)& 137 (MVCT)  \\
           &    &   &   &    &    &    &    & 95 (kVCT) & 98 (kVCT)  \\ \hline
  Slices & 16 & 32 & 64 & 128 & 256 & 512 & 1024 & 2048 & 2745(MVCT)   \\
  & & & & & & & & & 2824 (kVCT) \\ \hline
  Epochs & 5000 & 2500 & 1200 & 600 & 300 & 200 & 100 & 100 & 100 \\
  \hline \hline
  \end{tabular}
}
\end{table}

\Cref{Fig:histograms} shows the histograms of the images shown in \cref{Fig:VisualComp}.
The histograms of the processed MVCT images were similar to those of the kVCT images, regardless of the size of the training dataset.
The histograms of the processed MVCT images did not necessarily match perfectly with those of the kVCT image as they were acquired on different days.
In the histograms of the processed MVCT images, the peaks corresponding to the muscle and fat were well-separated.
This finding indicates that the contrast between the muscle and fat was enhanced in the processed MVCT images,
which is consistent with that obtained through the visual comparisons.

\begin{figure*}[t]
\includegraphics[width=0.5 \linewidth]{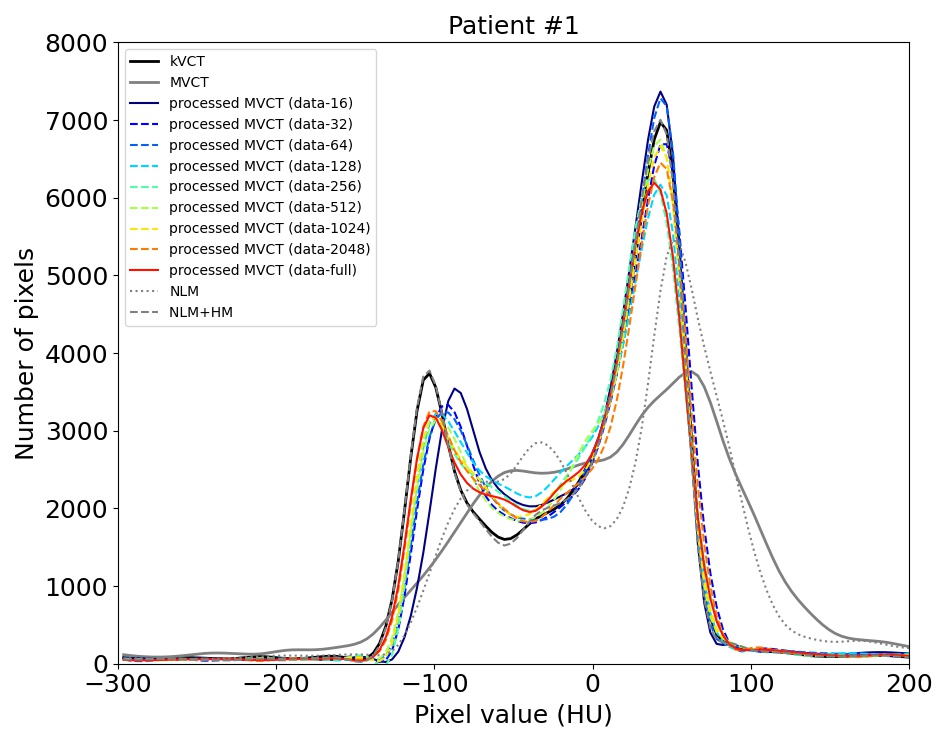}
\includegraphics[width=0.5 \linewidth]{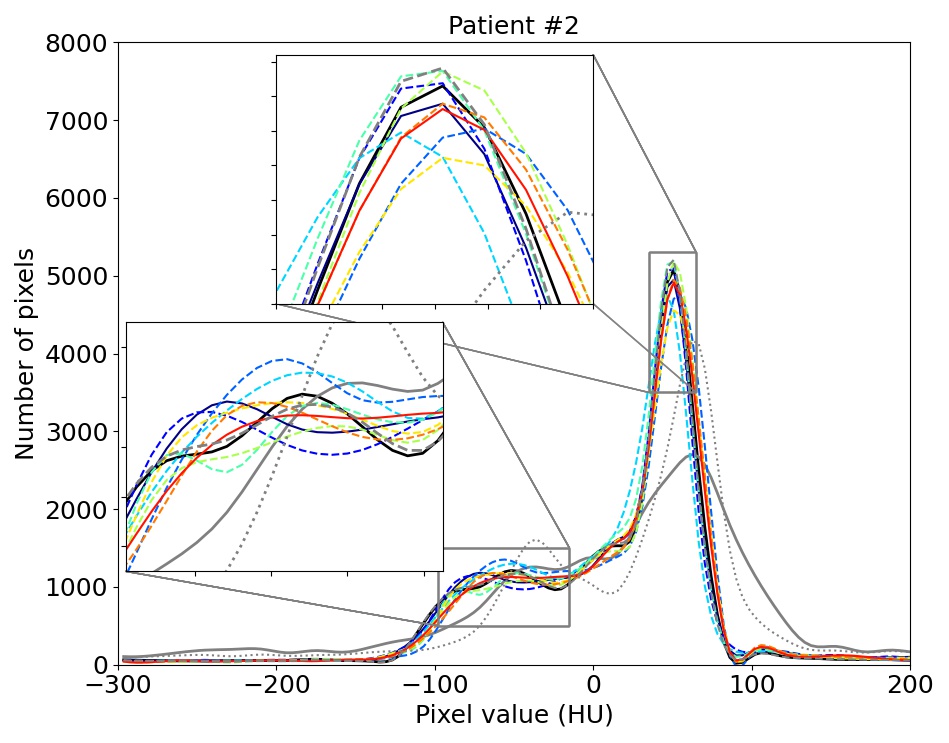}
\captionv{12}{}{
Histograms of HU values of the slices of 
patient \#1 (left) 
and patient \#2 (right)
shown in \cref{Fig:VisualComp}. 
The black line represents the histogram of kVCT, and the gray solid line represents the histogram of MVCT.
The gray dotted and dashed lines represent the histograms of NLM and NLM+HM, respectively.
\label{Fig:histograms}
}
\end{figure*}

\Cref{Fig:NSelfSSIM} shows the
distributions of NSelfSSIM defined in Eq~\eqref{Eq:NselfSSIM} for 
the validation slices. For the computation of NSelfSSIM, we used 
the $80\times 80$ square region around the center, 
which was inside the body for all 319 validation slices.
The HU values were normalized to [0, 255] before computing NSelfSSIM. 
The processed MVCT by the proposed model and original CycleGAN,
regardless of the training data size,
exhibit a low NSelfSSIM similar to that of kVCT, indicating the reduction of noise.
It should be noted that NSelfSSIM is small, for an extreme example, when the images are just blank.
Hence, a good balance between structure preservation and noise reduction is important.
\Cref{Fig:NSelfSSIM} assures that the noise level in terms of NSelfSSIM
was similar between kVCT and processed MVCT, 
regardless of the model and data size,
and considerably lower than that of the original MVCT.

\begin{figure*}[t]
\centering
\includegraphics[width=0.65 \textwidth]{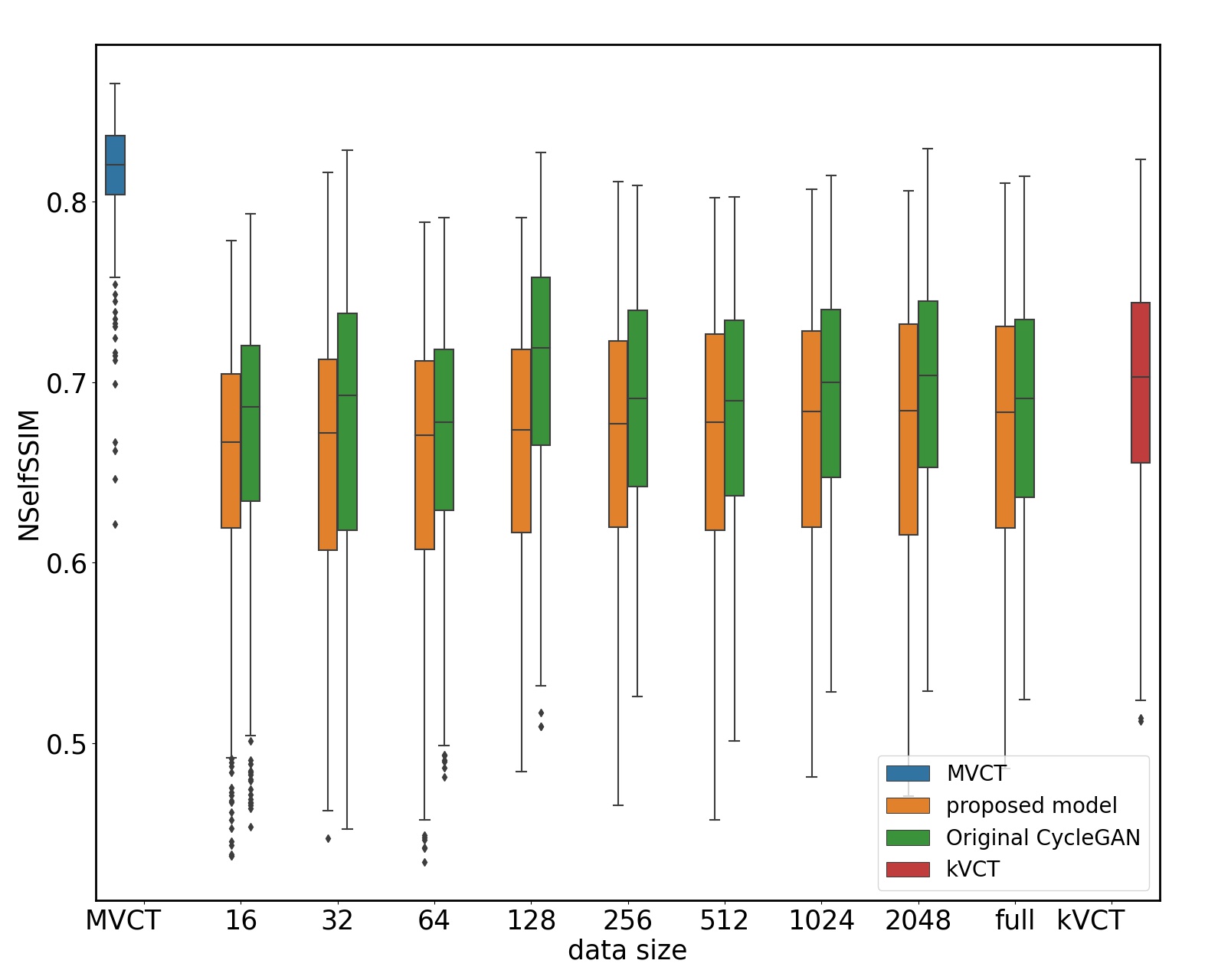}
\captionv{15}{}{
Distribution of NSelfSSIM for 319 validation slices from 16 patients.
Smaller values indicate that the images are less noisy.
The distributions for the original MVCT, processed MVCT by our model, and original CycleGAN
trained by datasets of various size are shown.
The distribution of kVCT images for 488 validation slices of the same patients but acquired on different days are also shown for reference.
\label{Fig:NSelfSSIM}
}
\end{figure*}

The box plots of the HU value distributions of soft tissues are shown in \cref{Fig:SoftTissues}.
Four ROIs of $10 \times 10$ pixels were extracted from each of patients \#1--\#4 for the fat, 
patients \#5--\#8 for the muscle,
patients \#9--\#12 for the spinal cord,
and patients \#13--\#16 for the tongue.
Accordingly, 16 ROIs were selected for each soft tissue to evaluate the distribution of the mean HU values.
The distributions of the 16 mean HU values of the ROIs were plotted for the original MVCT, processed MVCT by the proposed model, and original CycleGAN
trained using datasets of various sizes.
The distributions of the kVCT images for the same patients, acquired on different days, as well as the NLM+HM processed images (with the histogram matched to the kVCT after denoising)
are also shown for reference. 
The distributions of the processed MVCT by the proposed model and the original CycleGAN approach with respect to those of the kVCT as the dataset size increases can be observed.
Friedman's test~\cite{Friedman} (a non-parametric variant of parametric repeated measures ANOVA) along with a post hoc Nemenyi's non-parametric many-to-one comparison test~\cite{Nemenyi} 
were performed to compare the proposed models
trained with data-full and data-$n$ ($n\in \{16,32,64,128,256,512,1024,2048\})$.

\begin{figure*}[t]
\centering
\includegraphics[width=1.05 \textwidth]{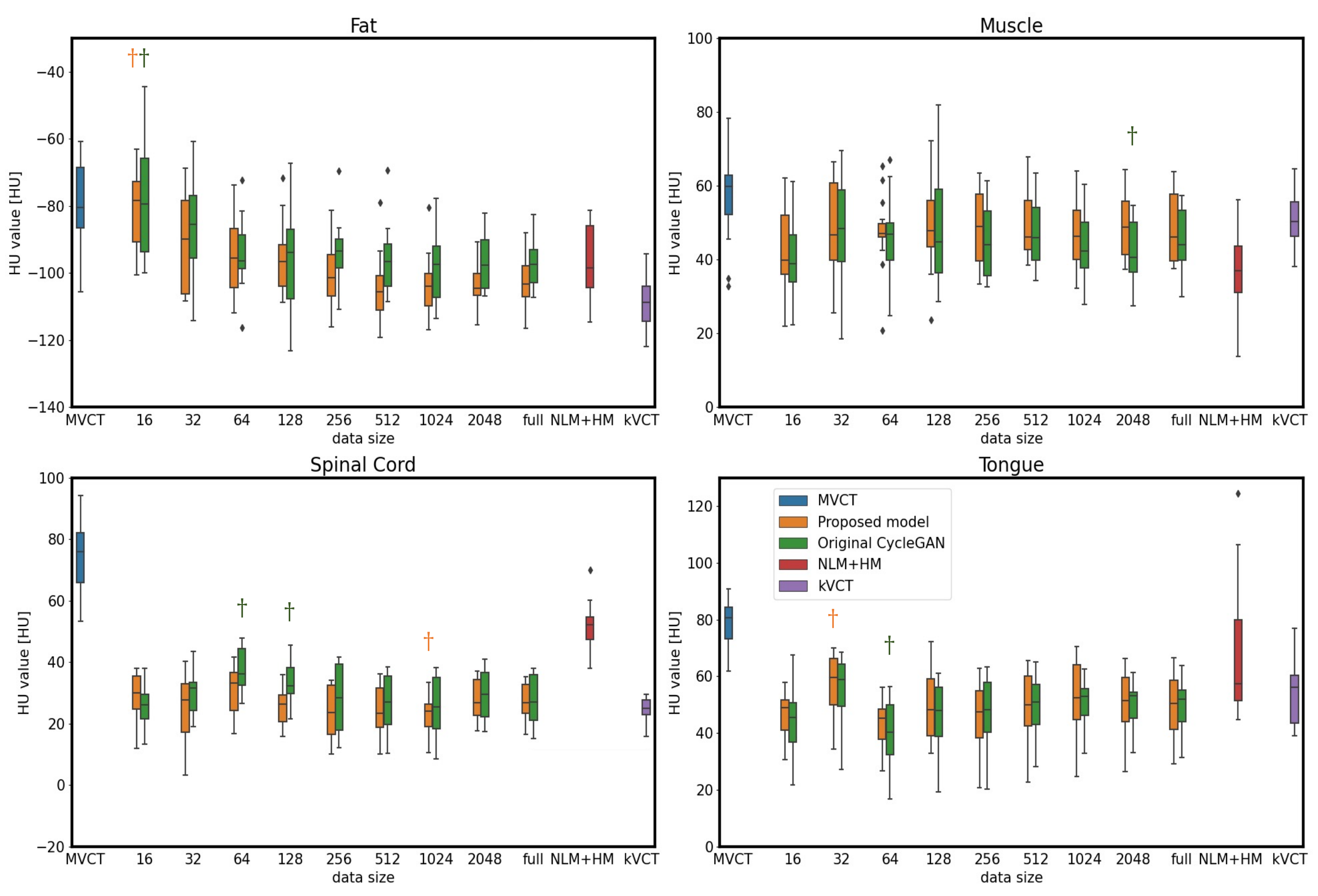}
\captionv{15}{}{
Distribution of the HU values of the 16 validation ROIs
selected from each soft tissue, including the fat, muscle, spinal cord, and tongue 
regions
 processed by
the proposed model and the original CycleGAN
trained with datasets of different sizes.
Distributions for MVCT and kVCT are also shown for reference.
Statistical analysis was performed using Friedman's test ($p < 0.05$) with a post hoc Nemenyi's test.
The dagger $\dag$ indicates $p < 0.05$ in Nemenyi's test.
\label{Fig:SoftTissues}
}
\vspace{5mm}
\end{figure*}

\Cref{Fig:heatmap} compares the DIG defined by Eq.~\eqref{Eq:DIG} of the images processed using the proposed and the original CycleGAN models trained with data-256. 
The high values of DIG represent the pixels where the edges are not aligned
between the original and processed MVCT.
We see that the edges of different structures and the body contour 
are better preserved with the proposed model (right panel)
than the original CycleGAN with a reduced weight of the adversarial loss (middle panel,
see also \cref{eq:originalCycleGAN} and the related discussion),
and the original CycleGAN (left panel).

Figure~\ref{Fig:DIG_datadep} shows the data size dependence of the DIG pertaining to the proposed model and the original CycleGAN.
The distributions of the mean of each of the 16 patients are plotted.
The difference between the proposed model and the original CycleGAN is 
particularly clear when the training data size is small.
Furthermore, to observe the convergence in terms of structure preservation,
Friedman's test along with a post hoc Shirley's test~\cite{Shirley} (a non-parametric variant of William's test)
were performed to compare the DIG for the proposed models
trained with data-$n$ and data-full ($n\in \{16,32,64,128,256,512,1024,2048\})$. 
We observed that the DIG of data-full did not differ significantly 
for $n\ge 256$ with the proposed model, whereas 
it did for $n=256$ with the original CycleGAN.
This suggests that the proposed model has better convergence 
with smaller data.

\begin{figure*}[t]
\centering
\includegraphics[width=1.05 \textwidth]{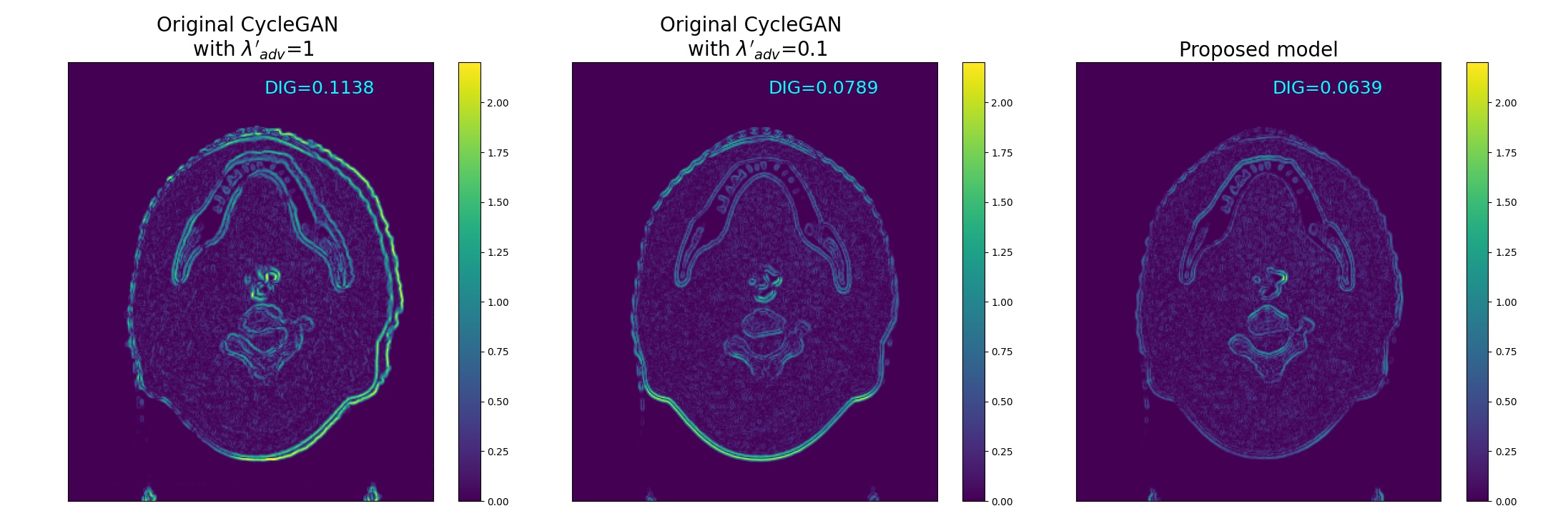}
\captionv{12}{}{
Visualization of DIG (Eq.~(\ref{Eq:DIG}))
for a selected slice from patient \#1 processed by 
the original CycleGAN with $\lambda'_\textrm{adv}=1.0$ (left panel),
the original CycleGAN with $\lambda'_\textrm{adv}=0.1$ (middle panel),
and the proposed model (right panel). See also \cref{eq:originalCycleGAN}.
All models were trained with data-256.
High values indicate poor low edge alignment
between the original and processed images.
\label{Fig:heatmap}
}
\end{figure*}

\begin{figure*}[t]
\centering
\includegraphics[width=0.7 \textwidth]{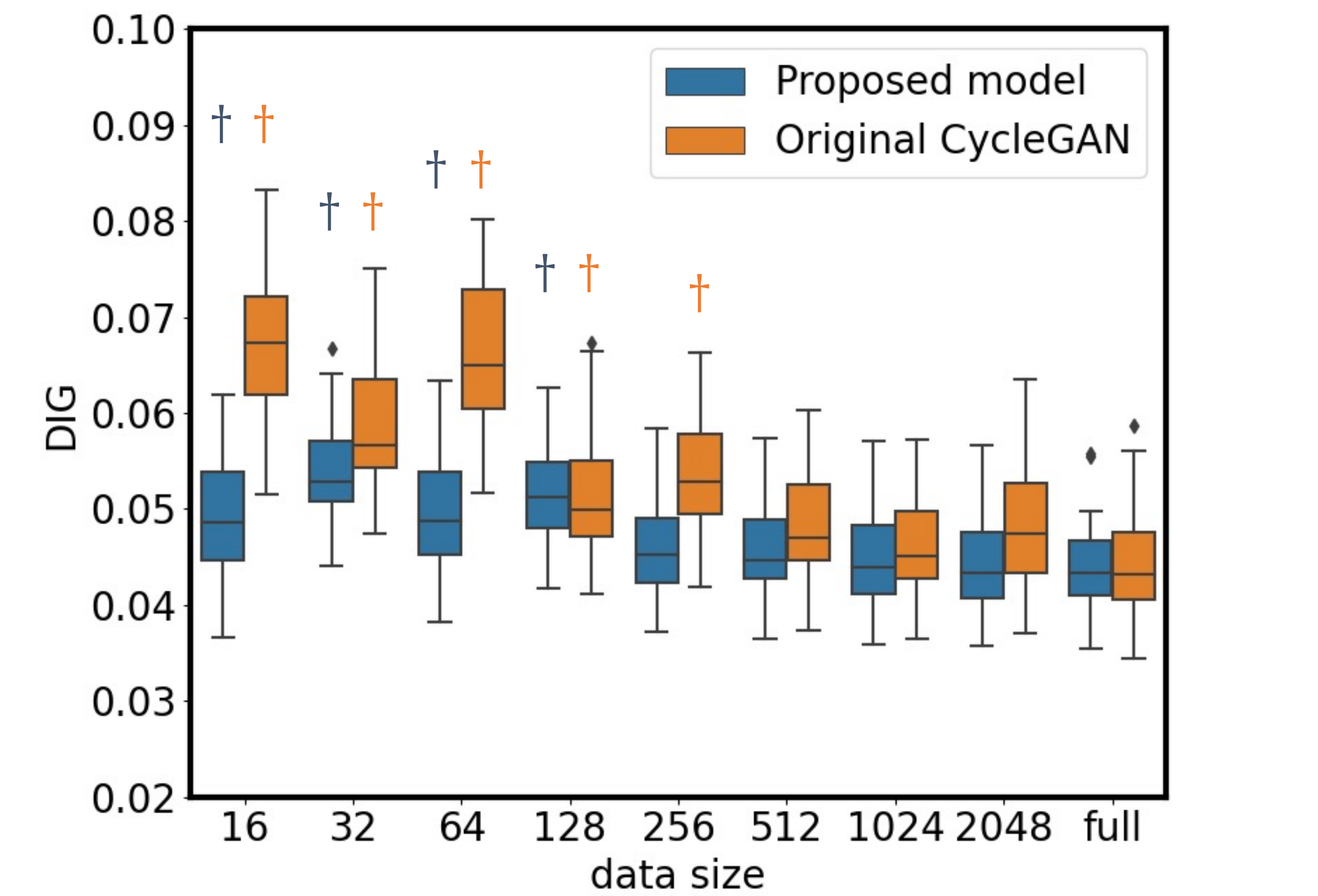}
\captionv{15}{}{
Performance comparison of models  
trained with datasets of different sizes
in terms of the structure preservation measured by DIG.
The box plot was produced for the distribution of the mean
of each patient among all the 16 patients.
Statistical analysis was performed by Friedman's test ($p < 0.05$) with a post hoc Shirley's test.
The dagger $\dag$ indicates $p < 0.05$ in Shirley's test.
\label{Fig:DIG_datadep}
}
\end{figure*}

In the field of modality conversions,
certain researchers reported the use of thousands of images~\cite{Wolterink:2017, Maspero:2018, kida:2019a, Liang:2019, Taasti:2020, Vinas:2021}.
The aforementioned results show that
the proposed model was successfully trained with 
a significantly smaller number of 256 slices.
This was confirmed by the convergence in the histogram of HU values as well as the two metrics to evaluate noise reduction and structure preservation.

\subsection{Effect of loss terms and data augmentation}

To investigate the role of each loss function, particularly in the preservation of structures, we present a comparison of the DIG among the original CycleGAN, original CycleGAN with total variation regularization, original CycleGAN with autoencoder loss, original CycleGAN with air loss, original CycleGAN with perceptual loss, and proposed model in \cref{Fig:DIG_losses}. 
Each hyperparameter is set to be equal to Eq.~(\ref{Eq:Hyperparametes}).
In this analysis, data-256 was used for training, and all the validation images were employed.
We observed that the autoencoder, air, and perceptual losses contributed to the preservation of the structures, as expected.
\Cref{Fig:Noise_effects} shows a slice of the validation data
processed using the proposed model trained with and without the 
noise injection described in \cref{sec:model}.
The image
processed using the proposed model trained with the 
noise injection
did not exhibit uneven artifacts that were visible in that without the noise injection.
Thus, the noise injection contributes toward robust and stable conversion,
 as in the denoising autoencoder.

\begin{figure*}[t]
\includegraphics[width=0.8 \textwidth]{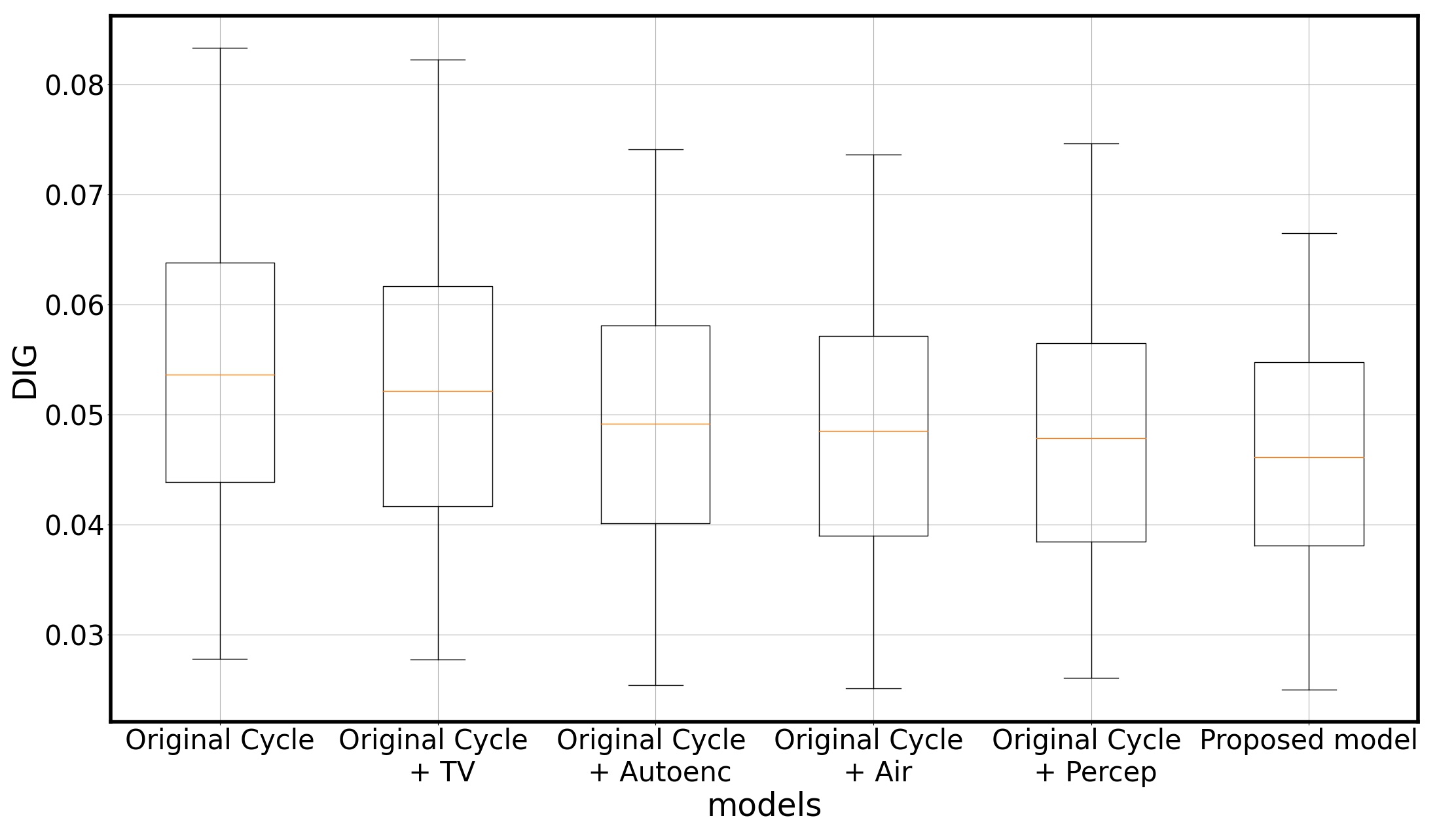}
\captionv{15}{}{
Performance comparison of the models with different loss functions in terms of the structure preservation measured by DIG.
The box plot was obtained in the same manner as for Fig.~\ref{Fig:DIG_datadep}.
From left to right, the results of the original CycleGAN, original CycleGAN with total variation (TV), original CycleGAN with autoencoder loss, original CycleGAN with air loss, original CycleGAN with perceptual loss, and proposed model are shown. In this analysis, data-256 was used for training.
\label{Fig:DIG_losses}
}
\end{figure*}

\begin{figure*}[t]
\includegraphics[width=1.0 \textwidth]{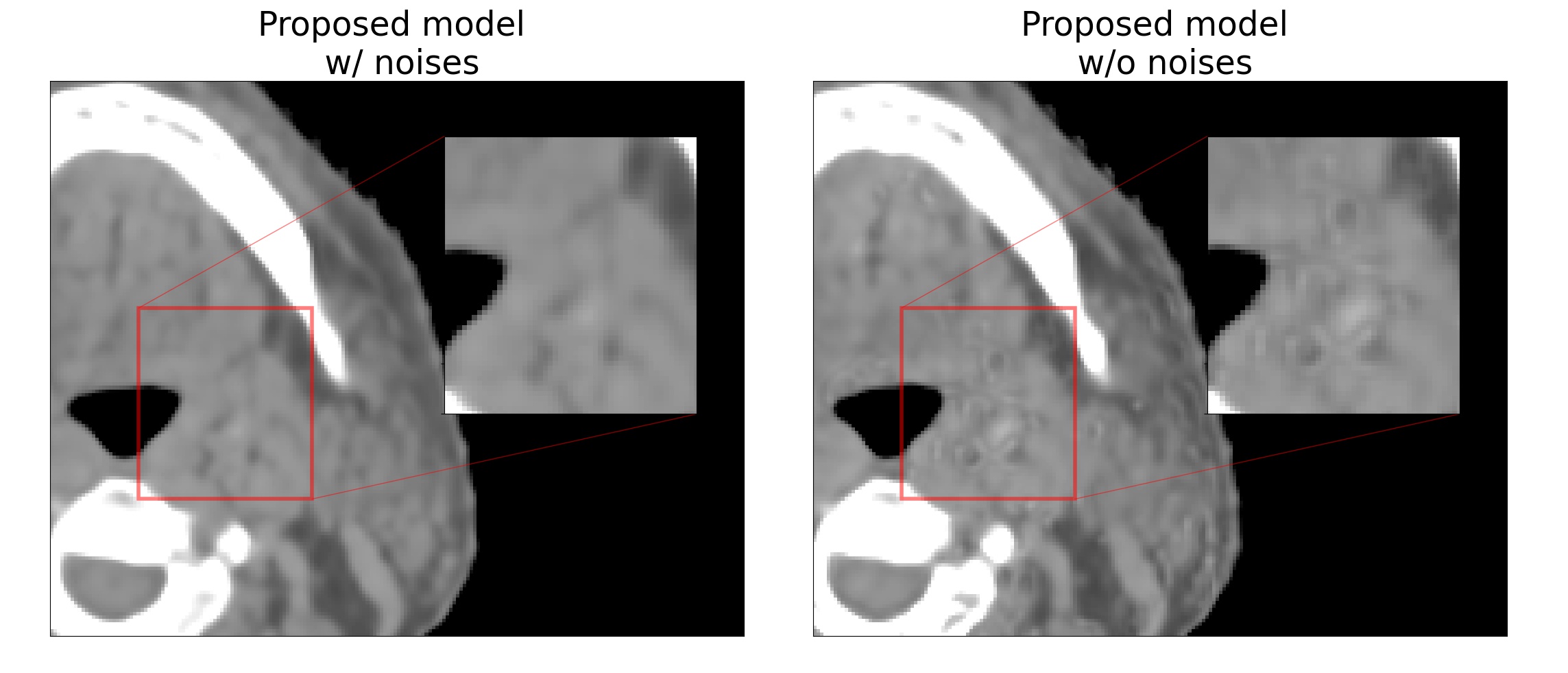}
\captionv{15}{}{
Uneven artifacts in the processed images.
The artifacts are suppressed
by the noise induction in the image and latent domains.
\label{Fig:Noise_effects}
}
\end{figure*}

\subsection{Clinical evaluation by contouring}

\Cref{Fig:DICE} shows the results of
contouring described in \cref{sec:contour}
in terms of the Dice coefficients for the 56 contours of the parotid glands
in each image modality.
We performed Friedman's test together with a post hoc Nemenyi's test.
In Friedman's test, we found a statistically significant difference among the Dice coefficients between the ground truth contouring and the contouring on MVCT, the processed MVCT by proposed models trained with data-16, data-256, and data-full.
Thus, a post hoc Nemenyi's test was performed to compare data-16, which had the smallest median,
and the other three cases.
The median values of data-256 and data-full were larger than that of data-16, whereas 
the difference between data-16 and MVCT was not significant.
This implies that the inter-person variability was lower 
with data-256 and data-full than MVCT and data-16,
indicating the improvement in the precision of contouring.
This finding also supports the aforementioned quantitative evaluation
by the metrics, where the performance of 
the model was noted to converge at the training data size of 256 slices.

\begin{figure*}[t]
\centering
\includegraphics[width=0.8\linewidth]{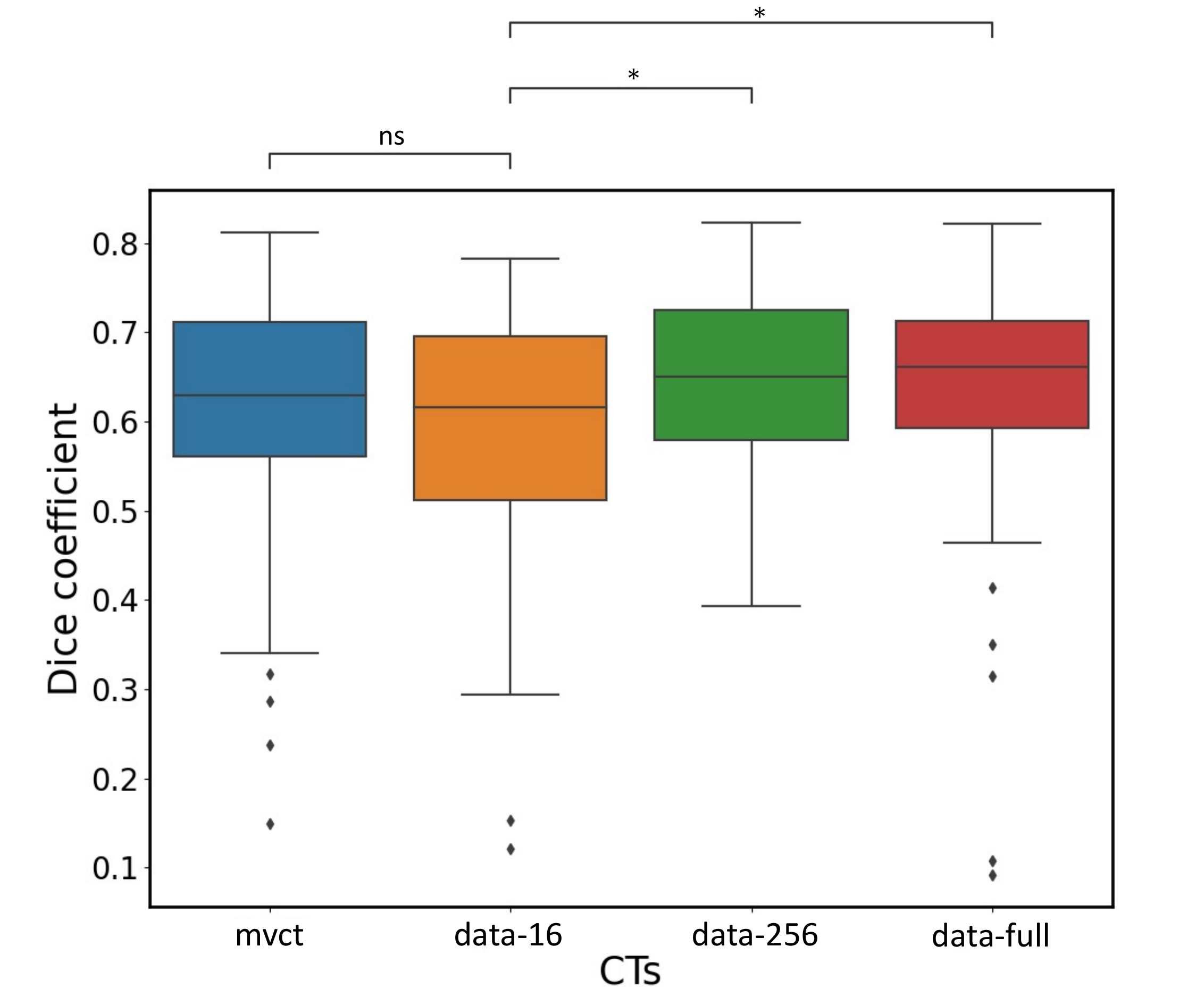}
\captionv{12}{}{Distribution of the Dice coefficients between the ground truth 
contouring 
of the parotid glands of the four patients in 
the validation data
and the contouring by seven doctors on 
MVCT, the processed MVCT
by the proposed models trained with data-16, data-256, and data-full.
Eventually, 56 contours per image modality of the parotid gland were used.
Statistical analysis was performed using Friedman's test ($p < 0.05$) with a post hoc Nemenyi's test for
comparing data-16 with the other three cases.
The asterisk $*$ indicates $p < 0.05$ in Nemenyi's test,
while ``ns'' indicates no significant difference.
\label{Fig:DICE} }
\end{figure*}


\section{Discussion}\label{sec:discussion}

\subsection{GAN-based modality conversion}
The primary purpose of MVCT to kVCT conversion is to extract as much information as possible from an MVCT image that appears deteriorated to the naked human eye. 
For example, some hardly visible structure in the MVCT image was made clear in the 
converted kVCT-like image (\cref{Fig:VisualComp}).
To achieve this conversion, in addition to hand-crafted priors such as the total variation,
GANs were used to obtain a prior for the distribution of kVCT images
in a data-driven manner. 
As GANs can potentially alter and forge 
structures that appear real, 
specific attention was paid to the preservation of anatomical structures
in addition to image quality enhancement (\cref{sec:technique}).
This is also consistent with the practical demand, as MVCT is widely used for image registration. 
The best method to evaluate a modality conversion method 
is to compare the processed images with the ground truth, 
which in our case refers to the kVCT images with the same structures as those in the MVCT images.
However, it is virtually impossible to obtain
MVCT and kVCT images simultaneously.
The evaluation is considerably difficult in the absence of the ground truth.
Consequently, we presented two
metrics, DIG and NSelfSSIM, 
that quantified the structure preservation and the noise level
in the processed image, respectively (\cref{sec:metrics}).

\subsection{Data size required for training}
Deep learning-based methods depend on
large amounts of training data; however,
the amount of data necessary for a particular task is often unclear.
Most previous studies do not provide an argument regarding the determination of the data size.
In this study, our primary goal was to reduce the size of the training dataset. We evaluated 
the model performance
with respect to the number of slices in the training dataset
in terms of
the distribution of HU values (\cref{Fig:histograms}),
noise reduction (\cref{Fig:NSelfSSIM}),
and structure preservation (\cref{Fig:DIG_datadep}).
As the number of slices in the training data
was increased from a few hundred to the thousands,
the metrics converged, as expected.
We observed convergence
at a number as low as 256 slices (from 32 patients).
This approach of observing the convergence 
of model performance generally provides a criterion
for a judicious choice of the data size (not restricted to specifically our case).

\subsection{Usability in a clinical task}
In deep learning, particularly when a GAN is
employed,
the resulting images should be used with extreme care 
in clinical tasks.
Their validity must be confirmed for each usage case.
As MVCT is widely used for image registration,
it would be beneficial if we could prove that 
MVCT processed with the proposed model 
improves the registration accuracy.
However, it is difficult to verify this directly
owing to the lack of ground truth.
As image registration relies largely on the 
distinguishability of different regions,
we evaluated a contouring task,
where the ground truth was defined by the average of 
the contouring performed by seven medical doctors.
We confirmed that the inter-person variability 
was lower with the processed MVCT (\cref{Fig:DICE}),
confirming the improvement in precision.
This result suggests that 
the proposed method would facilitate an improvement in the 
distinguishability of different regions.
Moreover, the enhanced contours in the processed MVCT
could benefit ART owing to the improved accuracy of the contouring of the target region and organs at risk.

\subsection{Generalizability of the proposed model}
Here, we discuss the possibilities of the generalization of the proposed model.
Our training dataset comprised images from the mandible to the base of the neck, avoiding the metal artifact in the oral cavity. 
We tested the proposed model by converting MVCT images to kVCT-like images
that were different from the ones used for the training; our test images included MVCT from
the head and pelvic regions and the oral cavity region with metal artifacts.
Figure~\ref{Fig:TransferLearning1} shows the MVCT, processed MVCT by the data-256 model, and reference kVCT of
 the head, pelvic, and metal artifact regions.
The images processed with the proposed model
were found to be fairly clear and accurate.
In the presence of metal artifacts, MVCT processed by the proposed model would be better suited for image registration than kVCT,
which was severely distorted by the artifacts.
The degree to which the 
processed images are usable for clinical tasks
remains to be carefully investigated.
Moreover, the applicability of transfer learning 
from one region to another would be an interesting 
future research topic.

\begin{figure*}[t]
\centering
\includegraphics[width=0.7 \textwidth]{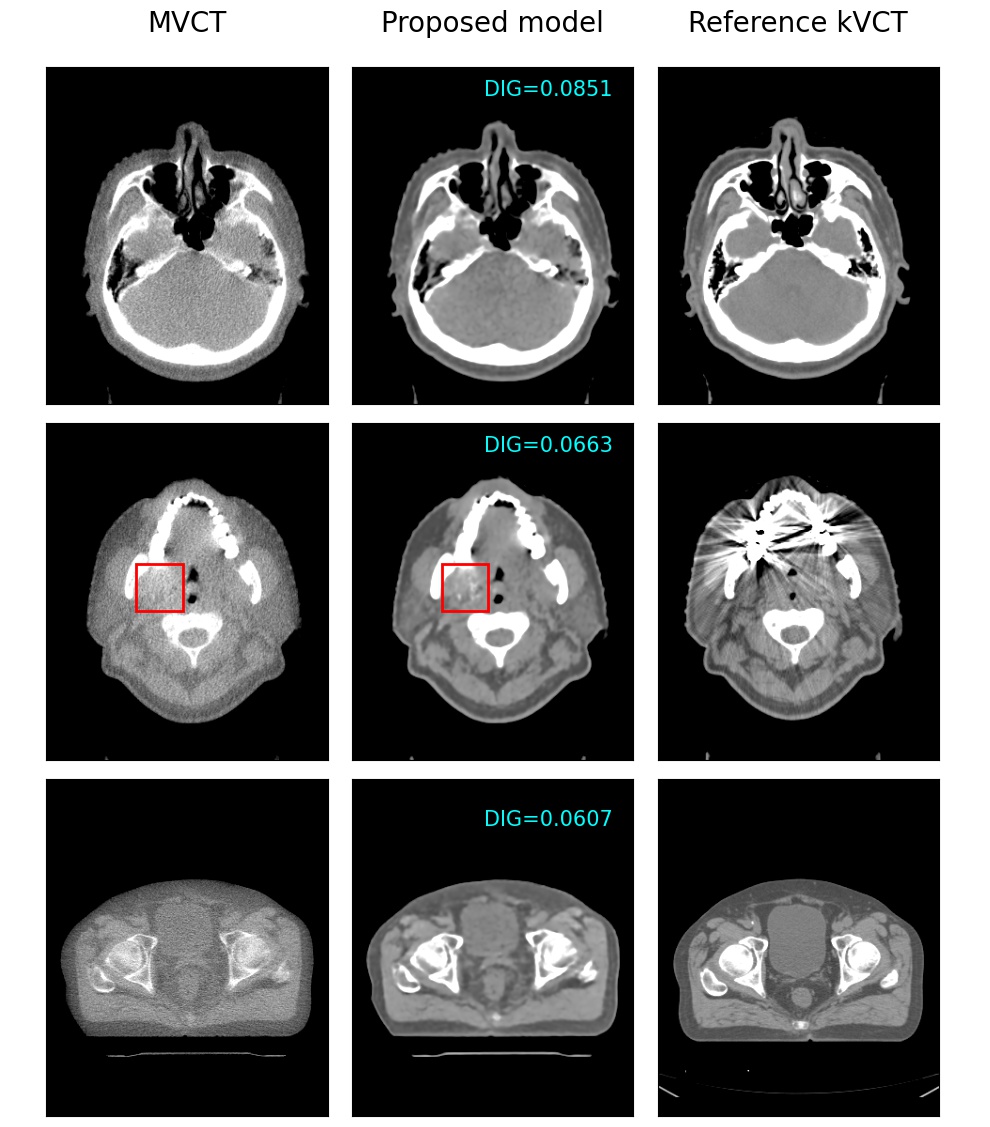}
\captionv{15}{}{
Generalizability of the proposed model.
The head region (first row), metal artifact region (second row), and pelvic region (third row) are shown.
The metal artifact in the second row is marked by the red box.
The display window range is set to (-300, 300) HU for all images.
\label{Fig:TransferLearning1}
}
\end{figure*}

\section{Conclusion}\label{sec:conclusion}

We constructed a modality conversion model based on deep learning and applied it to realize MVCT to kVCT conversion to enhance the image quality of MVCT used in helical tomotherapy.
The proposed model was designed with an emphasis on 
the structure preservation and 
size reduction of the training data,
which are among the most important factors 
for the application of 
deep-learning-based image processing in the medical field.
It was quantitatively shown that
the conversion 
model did not alter or forge structures that did not exist 
in the input image by comparing the
edges between different regions of input 
and processed images in the gradient domain.
We observed that training data with 256 slices from 32 patients, which
were significantly fewer than the number of slices used in previous works, were adequate.
As data collection is expensive in the medical field, this reduction is highly beneficial.

In addition to the visual and quantitative evaluations, 
we reported a clinical advantage of our method. 
In a contouring task, the doctors demonstrated
more consistent contouring on the processed MVCT
than on the original MVCT.
The training data size dependency was also evaluated 
with the same procedure,
confirming the aforementioned fact that
256 slices were sufficient.
The applicability of the processed images to clinical tasks other than contouring will be investigated in the future.

The proposed model exhibited some generalizability to 
process input images that were completely different from 
those in the training data.
The model converted the MVCT images containing metal artifacts to kVCT-like images reasonably well, 
even though the training data did not contain any images
with metal artifacts.
The MVCT images of the head and pelvic regions
were also successfully converted to kVCT images
by the model trained with images of the mandible to the neck regions.
Future work will be aimed at identifying the limitations concerning generalizability and transferability.

\section*{Acknowledgements}
\addcontentsline{toc}{section}{\numberline{}Acknowledgements}

Sho Ozaki was supported in part by the JSPS Grant-in-Aid for Scientific Research (C), 19K08093.
Kanabu Nawa was supported in part by the JSPS Grant-in-Aid for Scientific Research (C), 20K08073.
Toshikazu Imae was supported in part by the JSPS Grant-in-Aid for Scientific Research (C), 21K12121.
Takahiro Nakamoto was supported in part by the JSPS Grant-in-Aid for Young Scientists, 18K15625.
Akihiro Haga was supported in part by the JSPS Grant-in-Aid for Scientific Research (C), 19K08201.
Keiichi Nakagawa was supported in part by the JSPS Grant-in-Aid for Scientific Research (B), 20H04278.
We would like to thank Editage (www.editage.com) for English language editing.

\section*{References}
\addcontentsline{toc}{section}{\numberline{}References}
\vspace*{-20mm}






\bibliographystyle{./medphy.bst}    


\newpage

\end{document}